\documentclass[10pt,twocolumn,letterpaper]{article}

 \usepackage[pagenumbers]{cvpr} % To force page numbers, e.g. for an arXiv version

\usepackage[dvipsnames]{xcolor}

\definecolor{cvprblue}{rgb}{0.21,0.49,0.74}
\usepackage[pagebackref,breaklinks,colorlinks,citecolor=cvprblue]{hyperref}

\usepackage{multirow}
\usepackage{amssymb}

\usepackage{caption}
\usepackage{graphicx}
\usepackage{lipsum}

\usepackage{graphicx}
\usepackage{float}
\usepackage{overpic}

\def\paperID{5857} % *** Enter the Paper ID here
\def\confName{CVPR}
\def\confYear{2024}

\title{Frequency Domain Nuances Mining for Visible-Infrared Person Re-identification}

\author{Yukang Zhang\textsuperscript{\rm 1,2}, Yang Lu\textsuperscript{\rm 1,2}, Yan Yan\textsuperscript{\rm 1,2}, Hanzi Wang\textsuperscript{\rm 1,2}\thanks{Corresponding author.}, Xuelong Li\textsuperscript{\rm 3}\\
\textsuperscript{\rm 1}Fujian Key Laboratory of Sensing and Computing for Smart City, \\School of Informatics, Xiamen University, 361005, P.R. China.\\
\textsuperscript{\rm 2}Key Laboratory of Multimedia Trusted Perception and Efficient Computing, \\Ministry of Education of China, Xiamen University, 361005, P.R. China.\\
\textsuperscript{\rm 3}Northwestern Polytechnical University, Xi’an, Shaanxi, China.\\
{\tt\small zhangyk@stu.xmu.edu.cn, \{luyang, yanyan, hanzi.wang\}@xmu.edu.cn, li@nwpu.edu.cn}\and
}

\begin{document}
\maketitle
\let\oldtwocolumn\twocolumn
\renewcommand\twocolumn[1][]{%
    \oldtwocolumn[{#1}{
    \begin{center}
           \includegraphics[height=16.5cm,width=17.5cm]{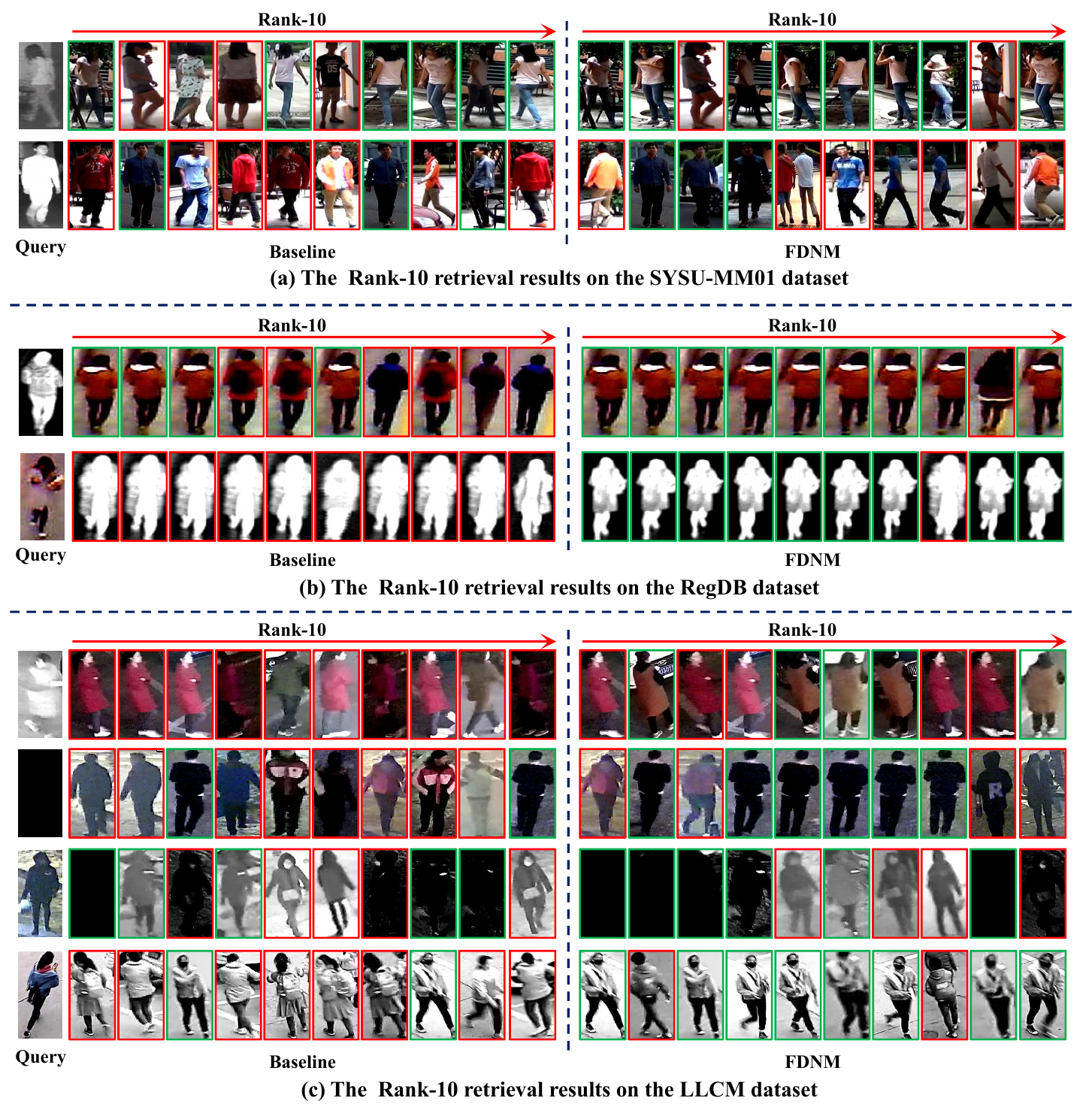}
           \captionof{figure}{The Rank-10 retrieval results obtained by the baseline method and the proposed FDNM method on the SYSU-MM01 (under the multi-shot setting and the all-search mode), RegDB and LLCM datasets. }
           \label{img8}
        \end{center}
    }]
}

\begin{abstract}
The key of visible-infrared person re-identification (VIReID) lies in how to minimize the modality discrepancy between visible and infrared images. Existing methods mainly exploit the spatial information while ignoring the discriminative frequency information. To address this issue, this paper aims to reduce the modality discrepancy from the frequency domain perspective. 
Specifically, we propose a novel Frequency Domain Nuances Mining (FDNM) method to explore the cross-modality frequency domain information, which mainly includes an amplitude guided phase (AGP) module and an amplitude nuances mining (ANM) module. These two modules are mutually beneficial to jointly explore frequency domain visible-infrared nuances, thereby effectively reducing the modality discrepancy in the frequency domain. Besides, we propose a center-guided nuances mining loss to encourage the ANM module to preserve discriminative identity information while discovering diverse cross-modality nuances. 
%To the best of our knowledge, this is the first work that explores the potential frequency information for VIReID research. 
Extensive experiments show that the proposed FDNM has significant advantages in improving the performance of VIReID. 
Specifically, our method outperforms the second-best method by 5.2\% in Rank-1 accuracy and 5.8\% in mAP on the SYSU-MM01 dataset under the indoor search mode, respectively. 
Besides, we also validate the effectiveness and generalization of our method on the challenging visible-infrared face recognition task. \textcolor{magenta}{The code will be available.}
\end{abstract}
    
%\vspace{-0.3cm}
\section{Introduction}
\label{sec:intro}

\begin{figure}
\centerline{\includegraphics[height=6.3cm,width=8.3cm]{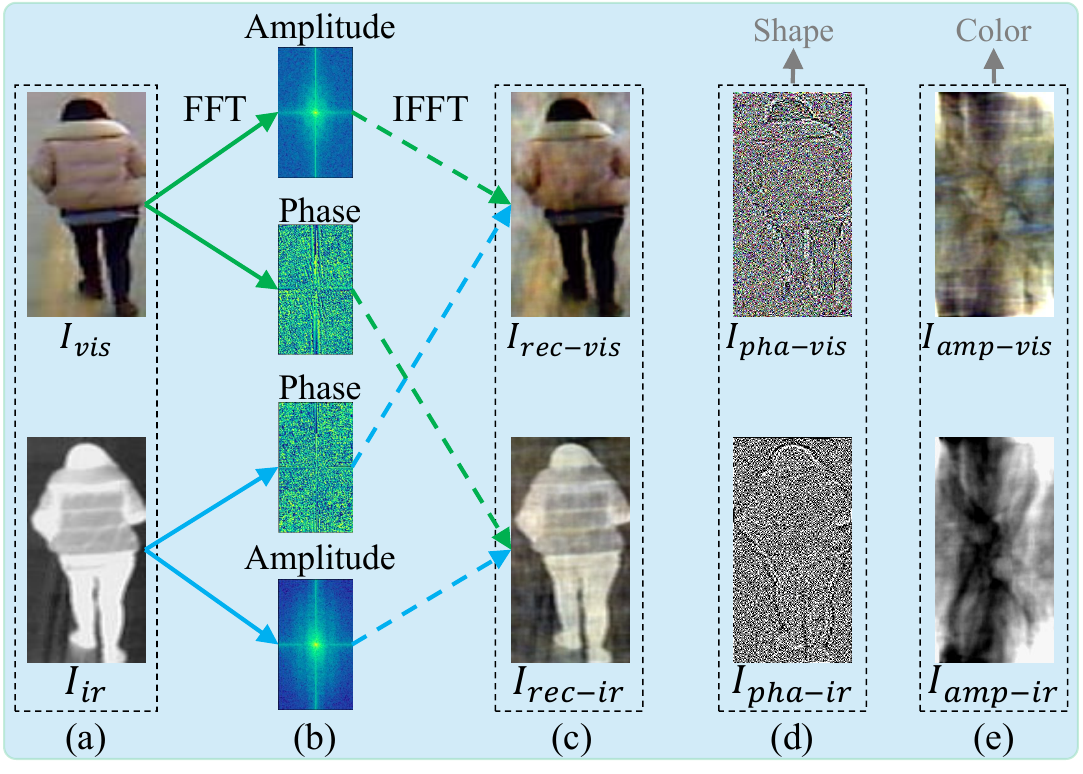}} \vspace{-0.23cm}
\caption{Decomposition and reconstruction of the VIS and IR images in the frequency domain. (a) denote the VIS and IR images; (b) are the amplitude and phase components of the VIS and IR images in the frequency domain; (c) are the reconstructed images of swapping the amplitude and phase components of the VIS and IR images; (d) are the reconstructed VIS and IR images with phase component information only; (e) are the reconstructed VIS and IR images with amplitude component information only. }%Decomposition and reconstruction of the VIS and IR images in the frequency domain. (a) denote the VIS and IR images; (b) are the amplitude and phase components of the VIS and IR images in the frequency domain; (c) are the reconstructed images of swapping the amplitude and phase components of the VIS and IR images; (d) are the reconstructed VIS and IR images with phase component information only; (e) are the reconstructed VIS and IR images with amplitude component information only.1}
\label{img1}
\vspace{-0.55cm}
\end{figure}

Visible-infrared (VIS-IR) person re-identification (VIReID) is attracting increasing attention due to its enormous potential in surveillance systems \cite{Shi_2023_ICCV}. Given a query image, the goal of VIReID is to match the VIS person images with corresponding IR person images in a non-overlapping camera surveillance system. The VIReID is challenging due to the large modality discrepancy between the VIS and IR images, resulting in significant intra-modality and inter-modality variations\cite{ye2018visible, ye2018hierarchical, lu2023learning}. 
Existing methods primarily focus on designing complex networks or loss functions to extract modality-invariant discriminative features \cite {hao2021cross, chen2021neural, cheng2023cross}. Although these methods have made remarkable progress in VIReID, they mainly exploit the spatial information but neglect the discriminative frequency information. 

\begin{figure}
\centerline{\includegraphics[height=4.3cm,width=8.3cm]{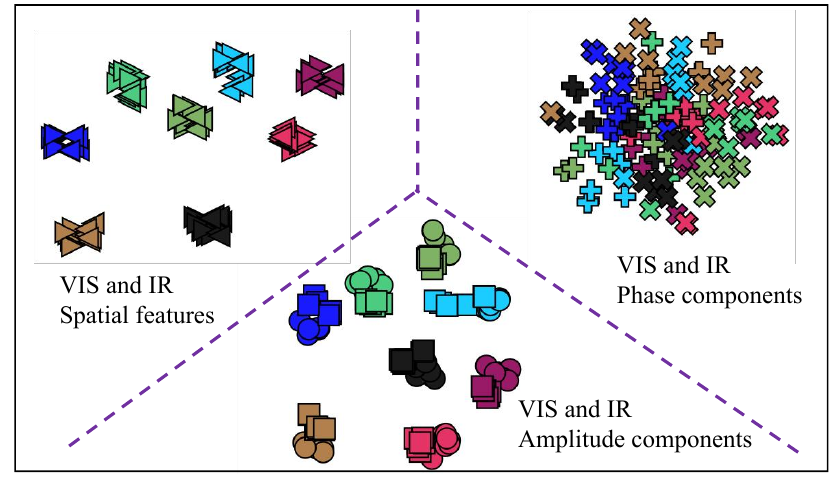}}\vspace{-0.15cm}
\caption{The t-SNE \cite{laurens2008Visualizing} for the VIS and IR features of spatial domain and frequency domain. The samples with the same color are from the same persons. It is apparent that the spatial features and the features of the amplitude component are discriminative while the features of the phase component contain key missing information for the VIReID task.}
\label{img2}
\vspace{-0.55cm}
\end{figure}

Recently, frequency domain feature learning has demonstrated its strong advantages in different tasks, such as image defogging\cite{yu2022frequency}, image exposure correction \cite{huang2022deep}, and domain generalization \cite{lee2023decompose}. Compared to the spatial domain information, the difference between the VIS and IR images in the frequency domain is physically defined. 
As shown in Fig. \ref{img1}, when reconstructing images by swapping the amplitude and phase components of the VIS and IR images, the resulting images have the same amplitude components but different phase components, making them visually similar. 
Besides, the images reconstructed by the phase components and the amplitude components preserve the shape and color information of person images, respectively. These results show that the amplitude components obtained by the Fourier transform contain key information for the VIReID task. Additionally, we also visualize the distribution of different features using t-SNE\cite{laurens2008Visualizing} in Fig. \ref{img2}. 
It is apparent that the distribution of the spatial features and amplitude components are distinguished, while the phase components contain important missing information. Thus, alleviating the modality discrepancy between the VIS and IR images and mining the cross-modality nuances in the frequency domain are of great importance for the VIReID task.

%Based on this observation, we propose a novel Frequency domain Nuances Mining Learning (FDNM) model to explore the cross-modality information in the frequency domain. 

Based on the above observation, we propose an effective frequency domain nuances mining (FDNM) method for the VIReID task. % to reduce the modality discrepancy between the VIS and IR images in the frequency domain. 
The proposed FDNM aims to enable the phase component to learn discriminative information and mine the diverse cross-modality nuances contained in the amplitude component. Therefore, we propose two key modules: an amplitude guided phase (AGP) module and an amplitude nuances mining (ANM) module. The proposed AGP module extracts the key information of the amplitude component to guide the learning of the phase component, and improve the discriminative ability of the phase component for cross-modality information. It enables the proposed method to effectively reduce the modality discrepancy between VIS and IR images in the frequency domain. The proposed ANM module is to fully explore the influence of cross-modality nuances in the amplitude component. To encourage the proposed ANM module to discover robust VIS-IR nuances, we propose an effective center-guided nuances mining loss, which can significantly improve the discriminative power of the proposed method. 
By combining the above two modules into an end-to-end network, the proposed FDNM can fully explore and utilize the diverse information in the frequency domain, significantly reducing the modality discrepancy between VIS and IR images. 
Extensive experiments show that the proposed FDNM achieves an impressive performance on three challenging VIReID datasets.  
The main contributions of this paper can be summarized as follows:

$\bullet$ We propose a novel frequency domain nuances mining method to exploits the potential frequency information for the VIReID task.

$\bullet$ We propose an amplitude guided phase module to exploit the key information of the amplitude component to promote the learning of the phase component, which improves the discriminative ability of the phase component and enables the proposed method to effectively learn robust VIS-IR feature representations.
%reduce the modality discrepancy between VIS and IR images.

$\bullet$ We propose an amplitude nuances mining module with a center-guided nuances mining loss to fully exploit the diverse cross-modality nuances
%in the amplitude component. By deeply mining the discriminative cross-modality nuances
contained in the amplitude component, which can significantly improve the performance of the proposed method.

$\bullet$ Extensive experiments show that the proposed method outperforms several state-of-the-art methods by a remarkable margin on three challenging VIReID datasets. %, including SYSU-MM01 \citep{wu2017rgb}, RegDB \citep{nguyen2017person} and LLCM \citep{zhang2023diverse}. 
Besides, we also validate the effectiveness and generalization of our method on the challenging VIS-IR face recognition task.

\section{Related Work}
\label{sec:formatting}

\begin{figure*}[t]
\centerline{\includegraphics[height=10.5cm,width=17.5cm]{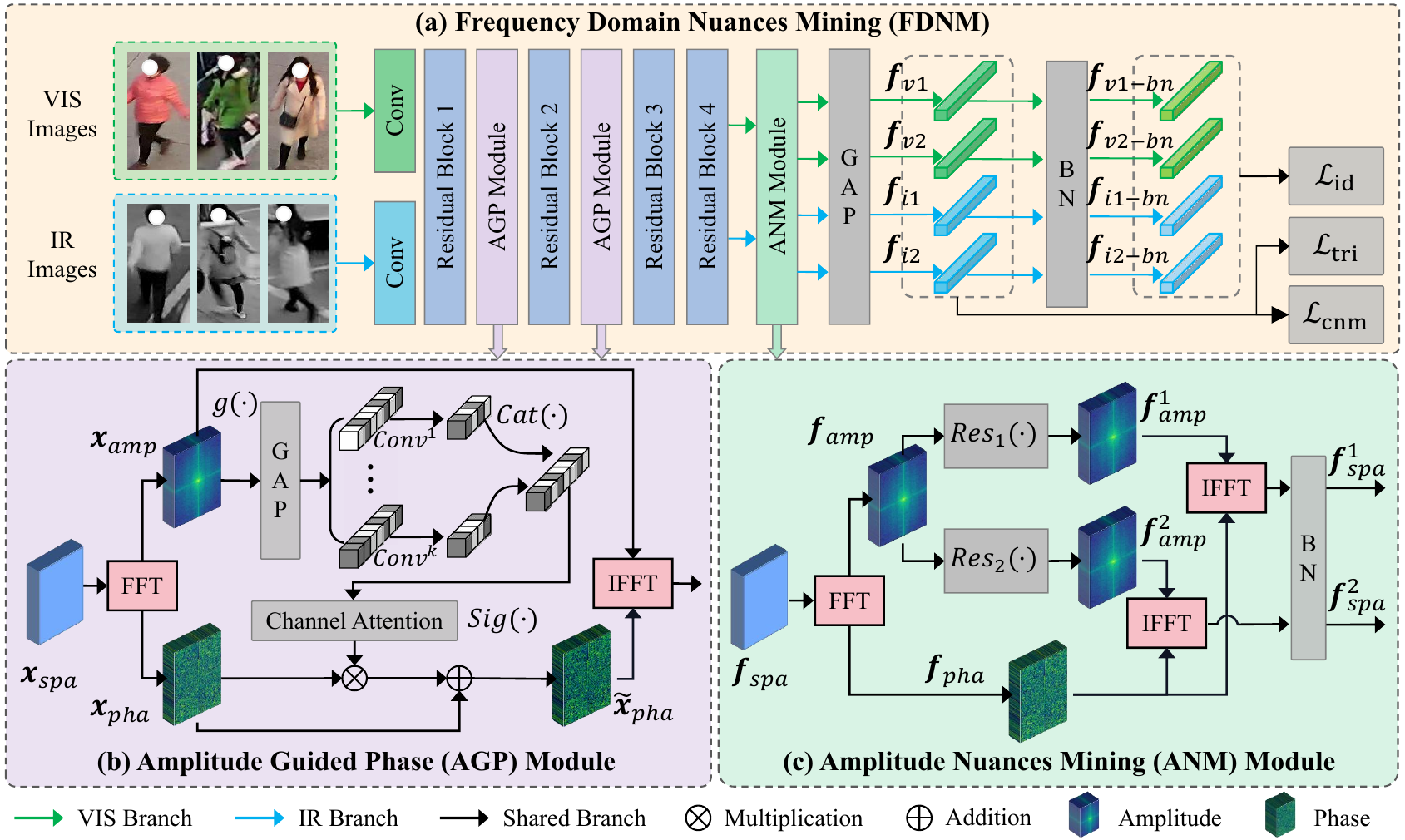}}\vspace{-0.2cm}
\caption{Overview of the proposed FDNM framework for the VIReID task. The proposed FDNM mainly includes an amplitude guided phase (AGP) module and an amplitude nuances mining (ANM) module to reduce the modality discrepancy between VIS and IR images in the frequency domain. After the GAP layer, we propose a center-guided nuances mining loss $\mathcal{L}_{cnm}$ to encourage the proposed ANM module to preserve discriminative identity information while discovering diverse cross-modality nuances. }
\label{img3}
\vspace{-0.35cm}
\end{figure*}

\subsection{Visible-Infrared Person Re-identification}

To reduce the VIS-IR modality discrepancy, numerous works have been proposed and achieved impressive performance \cite{ye2019modality, Wu_2023_ICCV}. Specifically, some image-level methods \cite{wang2019learning, wang2019aligngan, choi2020hi, wang2020cross} seek to generate pairs of the VIS-IR images for alleviating the modality gaps and improving the performance of the VIReID model. However, the lack of paired VIS-IR images may affect the quality and stability of the generated images. To address this issue, some lightweight networks \cite{li2020infrared, zhang2021towards, wei2021syncretic} apply a channel-level grayscale transformation to introduce another modality for assisting the VIS-IR match task. However, those methods cannot obtain consistent person poses of the VIS and IR images \cite{miao2021modality}. 

The other methods \cite{gao2021mso, pu2020dual, yang2023towards} try to find a latent feature space, where the VIS-IR modality discrepancy can be reduced. For instance, PartMix \cite{kim2023partmix} proposes to synthesize both positive and negative features and select reliable features to regularize the VIReID model. SGIEL \cite {Feng_2023_CVPR} proposes to jointly optimize shape-related and shape-erased objectives to learn modality-shared features. DEEN \cite{zhang2023diverse} generates diverse embeddings and mines the channel-wise and spatial embeddings to learn the informative features for reducing the VIS-IR modality discrepancy. SAAI \cite{Fang_2023_ICCV} proposes to align latent semantic parts by considering the similarity between pixel-level features and learnable prototypes. Although these methods have made impressive progress, they mainly exploit the spatial information but neglect the discriminative frequency information.

\subsection{Fourier Transform}

%The Fourier transform is a widely used technique for analyzing frequency content, and numerous studies have demonstrated its effectiveness for addressing different tasks. For example, Huang \emph{et al.} \cite{huang2022deep} introduce a Fourier-based exposure correction network to restore amplitude and phase representations, thereby improving lightness and refining structures progressively for the exposure correction task. Zhou \emph{et al.} \cite{zhou2022spatial} propose a spatial-frequency information integration network to explore the potential solution for pan-sharpening task. Yu \emph{et al.} \cite{yu2022frequency} propose a dual-guidance network for generating high-quality haze-free images by completely utilizing dual guidance for the haze degradation task. Wang \emph{et al.} \cite{Wang_2023_CVPR} propose a spatial-frequency mutual network to fuse frequency and spatial information for exploring the potential information for face super-resolution. Yao \emph{et al.} \cite{yao2023generalized} propose a frequency-based perturbation to capture the essential component of lightness for lightness adaptation task. 

The Fourier transform is widely used to analyze the frequency content, 
and numerous studies have demonstrated its effectiveness for addressing different tasks. For example, FECNet \cite{huang2022deep} introduces a Fourier-based exposure correction network to restore the amplitude and phase representations, thereby improving the lightness information for the exposure correction task. SFII \cite{zhou2022spatial} introduces a spatial-frequency information integration method to learn the effective information representation for the pan-sharpening task. FSDGN \cite{yu2022frequency} proposes a frequency-spatial dual-guidance network to explore frequency-spatial domain potentials for the haze degradation task. 
SFMNet \cite{Wang_2023_CVPR} designs a frequency-spatial interaction network to fuse frequency and spatial information and boost the face super-resolution performance. CSNorm \cite{yao2023generalized} designs an amplitude-based perturbation module to extract the essential information of lightness for the lightness adaptation task. FSI \cite{Liu_2023_ICCV} proposes to eliminate the various diffractions and reconstruct the textures in the frequency and spatial domains. FDMNet \cite{li2024frequency} proposes a novel frequency domain modality-invariant feature learning framework to reduce modality discrepancy from the frequency domain perspective.
Motivated by the success of these works, we propose a frequency domain nuances mining method for the VIReID task to exploit cross-modality nuances in the frequency domain.

\section{Methodology}

\subsection{Basics of Fast Fourier Transform}

The key of the VIReID task lies in minimizing the modality discrepancy between VIS and IR images. Previous works mainly exploit the spatial information while neglecting the discriminative frequency information, which is crucial for improving the performance of VIReID models. Therefore, before introducing the proposed FDNM, we first review the basics of the fast Fourier transform (FFT). Given a feature $\boldsymbol{x}\in\mathbb{R}^{C \times H \times W}$ output by  network, the FFT of the feature $\boldsymbol{x}$ can be expressed as: 
\begin{equation}
%\mathcal{F}(\boldsymbol{x})(u, v)=\frac{1}{\sqrt{H W}} \sum_{h=0}^{H-1} \sum_{w=0}^{W-1} \boldsymbol{x}(h, w) e^{-2 j \pi\left(\frac{h}{H} u+\frac{w}{W} v\right)},
\mathcal{F}(\boldsymbol{x})(u, v)=\frac{1}{\sqrt{H W}} \sum_{h=0}^{H-1} \sum_{w=0}^{W-1} \boldsymbol{x}(h, w) e^{-2 j \pi\left(\frac{h}{H} u+\frac{w}{W} v\right)},
\label{eq5.1}
\end{equation}
where $j$ is the imaginary unit. $u$ and $v$ represent the horizontal and vertical coordinates of the feature $\boldsymbol{x}$. $\mathcal{F(\boldsymbol{x})}$ represents the FFT of feature $\boldsymbol{x}$. Accordingly, $\mathcal{F}^{-1}(\boldsymbol{x})$ represents the inverse fast Fourier transform (IFFT) of the feature $\boldsymbol{x}$. In the proposed FDNM, the FFT and IFFT are computed separately on each channel of feature maps.

In the frequency domain, $\mathcal{F}(\boldsymbol{x})(u, v)$ can be represented by the amplitude component $\boldsymbol{x}_{amp}$ and the phase component $\boldsymbol{x}_{pha}$, which can be expressed as:
\begin{equation}
    \boldsymbol{x}_{amp}(u, v)=\sqrt{R^{2}(\boldsymbol{x})(u, v)+I^{2}(\boldsymbol{x})(u, v)},
    \label{eq5.2}
\end{equation}
\begin{equation}
    \boldsymbol{x}_{pha}(u, v)=\arctan \left(\frac{I(\boldsymbol{x})(u, v)}{R(\boldsymbol{x})(u, v)}\right),
    \label{eq5.3}
\end{equation}
where $R(\boldsymbol{x})$ and $I(\boldsymbol{x})$ are the real and imaginary part of $\mathcal{F(\boldsymbol{x})}$, respectively. 

In the frequency domain, the amplitude component $\boldsymbol{x}_{amp}$ is mainly affected by brightness, contrast and modality discrepancy. The phase component $\boldsymbol{x}_{pha}$ includes the structure information of the feature $\boldsymbol{x}$ \cite{yu2022frequency, zhou2022spatial}. Thanks to the FFT, the amplitude component $\boldsymbol{x}_{amp}$ and phase component $\boldsymbol{x}_{pha}$ can capture the global receptive field, which meets the demand of the VIReID task for effective long-distance dependency modeling and extracting modality-invariant feature representations. Additionally, it can be observed from Fig. \ref{img2} that the amplitude component $\boldsymbol{x}_{amp}$ and the phase component $\boldsymbol{x}_{pha}$ can capture different cross-modality features. Therefore, by modeling the amplitude component $\boldsymbol{x}_{amp}$ and the phase component $\boldsymbol{x}_{pha}$, the modality discrepancy between VIS and IR images can be effectively reduced, and thus helping improve the performance of the VIReID model.

\subsection{Overview}

Based on the above analysis, we propose an effective frequency domain nuances mining (FDNM) method. As shown in Fig. \ref{img3}, the proposed FDNM mainly includes an amplitude guided phase (AGP) module and an amplitude nuances mining (ANM) module to alleviate the modality discrepancy between VIS and IR images in the frequency domain. To improve the performance of the proposed FDNM, we combine the global and local feature representations for integrating discriminative information (for an intuitive display, only the global features are shown in Fig. \ref{img2}. The local features have similar losses and displays. The parts are set to 4. ). After the residual
block 4 and global average pooling (GAP), a batch normalization (BN) layer is applied to stabilize the training process \citep{Luo2019Bags}, which is parameter-shared among the two modalities. %Finally, the features before and after the BN layer are fed into different loss functions to jointly optimize the proposed FDNM method. 

\subsection{Amplitude Guided Phase (AGP) Module}

The proposed amplitude guided phase (AGP) module aims to utilize the key information in the amplitude component to guide the phase component for learning discriminative feature representations and reducing the modality discrepancy between the VIS and IR images.

To achieve this, as shown in Fig. \ref{img3} (b), 
the input features $\boldsymbol{x}_{spa}\in\mathbb{R}^{C \times H \times W}$ of the AGP module are first transformed into the frequency domain by FFT to obtain its amplitude components $\boldsymbol{x}_{amp}\in\mathbb{R}^{C \times H1 \times W1}$ and the phase components $\boldsymbol{x}_{pha}\in\mathbb{R}^{C \times H1 \times W1}$. Then, the amplitude components $\boldsymbol{x}_{amp}$ are fed into a global average pooling layer $g(\cdot)$ and K $1 \times 1$ convolutional layers $Conv^{K}(\cdot)=\{Conv^{k}(\cdot)\}_{k=1}^{K}$ with non-shared parameters to reduce its dimension to $C/K$ and extract robust amplitude components. Then, the $K$ amplitude components are concatenated to restore its dimension back to $C$ and fed into a sigmoid activation function $Sig(\cdot)$ to obtain the key amplitude information $\mathcal{S}(\boldsymbol{x}_{amp})$, which can be expressed as follows:
\begin{equation}
    \mathcal{S}(\boldsymbol{x}_{amp})=Sig(Cat(Conv^{K}(g(\boldsymbol{x}_{amp})))),
    \label{eq5.4}
\end{equation}
where $Cat(\cdot)$ denotes the concatenate operation.

As mentioned above, the amplitude components include key information for VIReID. Therefore, we utilize these key amplitude information $\mathcal{S}(\boldsymbol{x}_{amp})$ to guide the phase component for learning discriminative feature representations, as shown in Fig. \ref{img3} (b), which can be expressed as follows:
\begin{equation}
    \boldsymbol{\widetilde{x}}_{pha} = \mathcal{S}(\boldsymbol{x}_{amp}) \otimes \boldsymbol{x}_{pha} + \boldsymbol{x}_{pha}. 
\end{equation}

Finally, the amplitude components $\boldsymbol{x}_{amp}$ and the obtained guided phase components $\boldsymbol{\widetilde{x}}_{pha}$ are transformed back into the spatial domain features using the IFFT. By incorporating the proposed AGP module into the backbone network, we can effectively enhance the ability to extract discriminative features and alleviate the VIS-IR modality discrepancy in the frequency domain, ultimately leading to improved performance in the VIReID task.

\subsection{Amplitude Nuances Mining (ANM) Module}

As shown in Fig. \ref{img3} (c), the proposed amplitude nuances mining (ANM) module aims to discover the cross-modality nuances between the VIS and IR images in the frequency domain with a dual-branch structure. By mining diverse information contained in the amplitude component, the proposed ANM module can in turn promote the learning of the AGP module on the phase component, thereby effectively improving the performance of the VIReID task.

Specifically, the proposed ANM module first transforms the features $\boldsymbol{f}_{spa}$ output by the last stage of the backbone into the frequency domain by FFT to obtain its amplitude components $\boldsymbol{f}_{amp}$ and the phase components $\boldsymbol{f}_{pha}$. Then, two $1 \times 1$ convolutional blocks $Res_{1}(\cdot)$ and $Res_{2}(\cdot)$ are applied to learn diverse amplitude component nuances, with two Instance Normalization (IN) layers \cite{ulyanov2016instance} to alleviate the modality discrepancy, which can be expressed as follows:
\begin{equation}
    \boldsymbol{f}^{1}_{amp}=Res_{1}(\boldsymbol{f}_{amp}), \quad \boldsymbol{f}^{2}_{amp}=Res_{2}(\boldsymbol{f}_{amp}).
\end{equation}

Then, these two amplitude components $\boldsymbol{f}^{1}_{amp}$ and $\boldsymbol{f}^{2}_{amp}$ are combined with the phase components $\boldsymbol{f}_{pha}$ using IFFT to obtain two different spatial domain features, respectively. Finally, two parameter-shared batch normalization (BN) layers are applied to make the ANM module easier to optimize. Nevertheless, on the one hand, directly applying the IN layers to amplitude components $\boldsymbol{f}^{1}_{amp}$ and $\boldsymbol{f}^{2}_{amp}$ may damage some discriminative identity information, thereby adversely affecting the performance. On the other hand, such two convolutional branches cannot effectively mine the cross-modality nuances in the amplitude components. 

\begin{figure}
\centerline{\includegraphics[height=4.5cm,width=8.3cm]{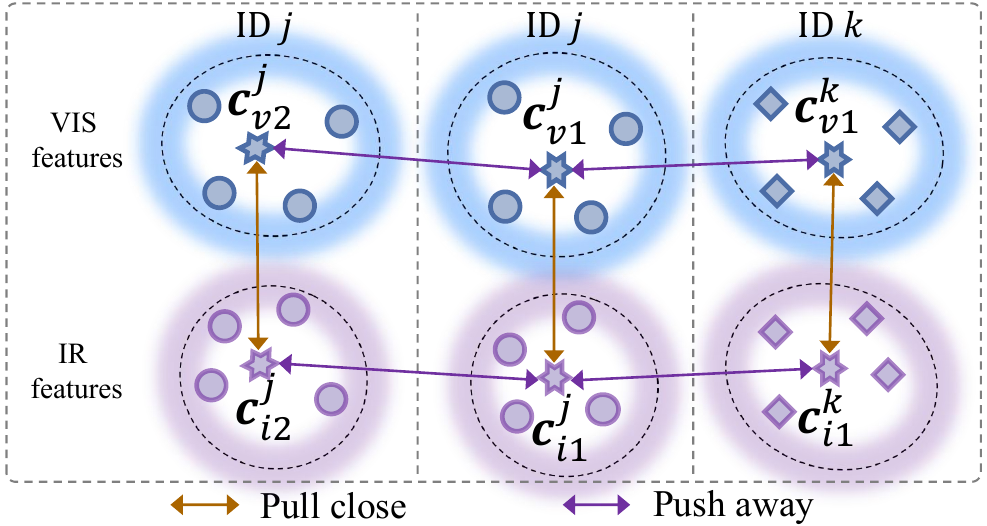}}\vspace{-0.2cm}
\caption{Illustration of the proposed center-guided nuances mining loss, which is used to preserve discriminative identity information while discovering diverse cross-modality nuances.}
\label{img4}
\vspace{-0.4cm}
\end{figure}

To address the above issues, we propose a center-guided nuances mining loss $\mathcal{L}_{cnm}$ to preserve discriminative identity information while discovering diverse cross-modality nuances. As shown in Fig. \ref{img4}, taking the class centers $\boldsymbol{c}_{v1}$ of feature $\boldsymbol{f}_{v1}$ from one VIS branch as an example, the center-guided nuances mining loss can be formulated as follows: 
\vspace{-0.15cm}
\begin{equation}
\begin{split}
\label{eq7}
\mathcal{L}_{cnm}(\boldsymbol{c}_{v1})
= \sum\limits_{\substack{j, k = 1,\\ y_j \neq y_k}}^P &[2 \times \textit{D}(\boldsymbol{c}^{j}_{v1}, \boldsymbol{c}^{j}_{i1}) 
- \textit{D}(\boldsymbol{c}^{j}_{v1}, \boldsymbol{c}^{j}_{v2}) \\
&- \textit{D}(\boldsymbol{c}^{j}_{v1}, \boldsymbol{c}^{k}_{v1}) + m]_{+}.
\end{split}
\end{equation}
where $\boldsymbol{\textit{D}}(\cdot, \cdot)$ represents the Euclidean distance between two features. $\boldsymbol{c}_{v1}$ and $\boldsymbol{c}_{i1}$ are the class centers of the VIS and IR features from the same branch in a minibatch, respectively. $\boldsymbol{c}_{v2}$ is the class centers of the VIS features from the other branch. $j$ and $k$ represent different classes in a minibatch. $m$ is a margin hyperparameter to balance the three terms. 

In Eq. (\ref{eq7}), $\mathcal{L}_{cnm}(\boldsymbol{c}_{v1})$ can achieve diverse amplitude nuances mining by forcing the distances $\textit{D}(\boldsymbol{c}^{j}_{v1}, \boldsymbol{c}^{j}_{v2})$ between the same class centers $\boldsymbol{c}^{j}_{v1}$ and $\boldsymbol{c}^{j}_{v2}$ in the two branches from the VIS modality to be larger than the distances $\textit{D}(\boldsymbol{c}^{j}_{v1}, \boldsymbol{c}^{j}_{i1}) $ between the class centers $\boldsymbol{c}^{j}_{v1}$ and $\boldsymbol{c}^{j}_{i1}$ of the VIS and IR modalities. Additionally, $\mathcal{L}_{cnm}(\boldsymbol{c}_{v1})$ can effectively alleviate the modality discrepancy between the VIS and IR features by forcing the distances $\textit{D}(\boldsymbol{c}^{j}_{v1}, \boldsymbol{c}^{j}_{i1}) $ between the class centers $\boldsymbol{c}^{j}_{v1}$ and $\boldsymbol{c}^{j}_{i1}$ from the same class but different modalities in the two branches to be smaller than the distances $\textit{D}(\boldsymbol{c}^{j}_{v1}, \boldsymbol{c}^{k}_{v1})$ between the feature centers $\boldsymbol{c}^{j}_{v1}$ and $\boldsymbol{c}^{k}_{v1}$ from different classes in the same modality.

Similarly, for the class centers $\boldsymbol{c}_{v2}$, $\boldsymbol{c}_{i1}$ and $\boldsymbol{c}_{i2}$ of different branches, we can get its loss: $\mathcal{L}_{cnm}(\boldsymbol{c}_{v2})$, $\mathcal{L}_{cnm}(\boldsymbol{c}_{i1})$ and $\mathcal{L}_{cnm}(\boldsymbol{c}_{i2})$. Therefore, the final center-guided nuances mining loss $\mathcal{L}_{cnm}$ can be formulated as follows: 
\vspace{-0.1cm}
\begin{equation}
\label{eq8}
\begin{split}
\mathcal{L}_{cnm} &= [ \mathcal{L}_{cnm}(\boldsymbol{c}_{v1}) + \mathcal{L}_{cnm}(\boldsymbol{c}_{v2}) \\ & + \mathcal{L}_{cnm}(\boldsymbol{c}_{i1}) + \mathcal{L}_{cnm}(\boldsymbol{c}_{i2})]/4.
\end{split}
\end{equation}

\subsection{Multi-Loss Optimization}

Besides the center-guided nuances mining loss $\mathcal{L}_{cnm}$, the cross-entropy loss $\mathcal{L}_{id}$ and the triplet loss \citep{hermans2017defense} $\mathcal{L}_{tri}$ are also used to jointly optimize the proposed FDNM. Following \citep{ye2021deep, zhang2023mrcn}, we adopt the cross-entropy loss $\mathcal{L}_{id}$ and the triplet loss $\mathcal{L}_{tri}$ as our baseline loss for learning discriminative features. Therefore, the overall loss ${\mathcal{L}_{fdnm}}$ for the proposed FDNM can be formulated as follows:
\begin{equation}
\label{eq9}
{\mathcal{L}_{fdnm}} = {\mathcal{L}_{id}} + \lambda_1 {\mathcal{L}_{tri}} + \lambda_2 {\mathcal{L}_{cnm}}, 
\end{equation}
where $\lambda_1$ and $\lambda_2$ are two hyperparameters to balance the relative importance of the loss terms ${\mathcal{L}_{tri}}$ and ${\mathcal{L}_{cnm}}$.
\section{Experiments}

%In this section, we compare the proposed FDNM method with several state-of-the-art methods, and also conduct the ablation studies to analyze the key components in the proposed FDNM method.

\begin{table*}%\footnotesize
  \centering
    \caption{Comparisons with several state-of-the-art methods on the SYSU-MM01 (single-shot setting) and RegDB datasets. R-1, R-10, R-20 denotes the Rank-1, Rank-10, Rank-20 accuracy, respectively. The best and the second results are \textcolor{red}{red} and \textcolor{blue}{blue} marked, respectively.
    }%Here, all the results obtained by the competing methods are measured using the single-query rule.}
    \label{tab1}
  \tabcolsep=0.14cm
   \vspace{-0.15cm}
  \renewcommand{\arraystretch}{1}
  \fontsize{9pt}{11pt}\selectfont
  \begin{tabular}{l||cccc|cccc||cccc|cccc}
  \hline  
   \multirow{3}{*}{Methods}  & \multicolumn{8}{c||}{SYSU-MM01} & \multicolumn{8}{c}{RegDB}\\ \cline{2-17} 
   &\multicolumn{4}{c|}{All Search} &\multicolumn{4}{c||}{Indoor Search}&\multicolumn{4}{c|}{VIS to IR}&\multicolumn{4}{c}{IR to VIS}\\
   \cline{2-17}
   & R-1 & R-10 & R-20 & mAP   & R-1 & R-10 & R-20 & mAP & R-1 & R-10 & R-20 & mAP & R-1 & R-10 & R-20 & mAP \\
   \hline 
    D$^{2}$RL \cite{wang2019learning} &  28.9 & 70.6 & 82.4 & 29.2 & -    & -    & -    & -       & 43.4 & 66.1 & 76.3 & 44.1 & -    & -    & -    & -       \\
    Hi-CMD \cite{choi2020hi}      &  34.9 & 77.6 & -    & 35.9 & -    & -    & -    & -       & 70.9 & 86.4 & -    & 66.0 & -    & -    & -    & -       \\
    %JSIA-ReID \cite{wang2020cross}  &  38.1 & 80.7 & 89.9 & 36.9 & 43.8 & 86.2 & 94.2 & 52.9    & 48.1 & -    & -    & 48.9 & 48.5 & -    & -    & 49.3    \\
    AlignGAN \cite{wang2019aligngan} &  42.4 & 85.0 & 93.7 & 40.7 & 45.9 & 87.6 & 94.4 & 54.3    & 57.9 & -    & -    & 53.6 & 56.3 & -    & -    & 53.4    \\
    cm-SSFT \citep{lu2020cross}       &  47.7 & -    & -    & 54.1 & -    & -    & -    & -      & 65.4 & -    & -    & 65.6 & 63.8 & -    & -    & 64.2    \\
    %X-Modality \cite{li2020infrared}  &  49.9 & 89.8 & 96.0 & 50.7 & -    & -    & -    & -       & 62.2 & 83.1 & 91.7 & 60.2 & -    & -    & -    & -       \\
    DDAG \cite{ye2020dynamic}       &  54.8 & 90.4 & 95.8 & 53.0 & 61.0 & 94.1 & 98.4 & 68.0   & 69.3 & 86.2 & 91.5 & 63.5 & 68.1 & 85.2 & 90.3 & 61.8    \\
    LbA \cite{park2021learning}      &  55.4 & -    & -    & 54.1 & 58.5 & -    & -    & 66.3   & 74.2 & -    & -    & 67.6 & 67.5 & -    & -    & 72.4    \\
    NFS \cite{chen2021neural}       &  56.9 & 91.3 & 96.5 & 55.5 & 62.8 & 96.5 & 99.1 & 69.8   & 80.5 & 91.6 & 95.1 & 72.1 & 78.0 & 90.5 & 93.6 & 69.8    \\
    CoAL \citep{wei2020co}         &  57.2 & 92.3 & 97.6 & 57.2 & 63.9 & 95.4 & 98.8 & 70.8   & 74.1 & 90.2 & 94.5 & 70.0 & -    & -    & -    & -       \\
    %DG-VAE \citep{pu2020dual}     &  59.5 & 93.8 & -    & 58.5 & -    & -    & -    & -      & 73.0 & 86.9 & -    & 71.8 & -    & -    & -    & -       \\
    CM-NAS \cite{fu2021cm}      &  60.8 & 92.1 & 96.8 & 58.9 & 68.0 & 94.8 & 97.9 & 52.4   & 82.8 & 95.1 & 97.7 & 79.3 & 81.7 & 94.1 & 96.9 & 77.6    \\
    %MSA \citep{miao2021modality}    &  63.1 & -    & -    & 59.2 & 67.2 & -    & -    & 72.7   & 84.9 & 92.8 & 95.1 & 82.2 & -    & -    & -    & -       \\
    MCLNet \cite{hao2021cross}   &  65.4 & 93.3 & 97.1 & 62.0 & 72.6 & 97.0 & 99.2 & 76.6   & 80.3 & 92.7 & 96.0 & 73.1 & 75.9 & 90.9 & 94.6 & 69.5    \\
    FMCNet \cite{Zhang_2022_CVPR}   &  66.3 & -    & -    & 62.5 & 68.2 & -    & - & 74.1      & 89.1 & -    & - & 84.4 & 88.4 & -    & - & 83.9 \\
    SMCL \cite{wei2021syncretic}   &  67.4 & 92.9 & 96.8 & 61.8 & 68.8 & 96.6 & 98.8 & 75.6    & 83.9 & -    & -    & 79.8 & 83.1 & -    & -    &78.6     \\
    PMT \citep{lu2023learning} &  67.5 & 95.4 & 98.6 & 65.0 & 71.7 & 96.7 & 99.3 & 76.5 & 84.8  & -& -& 76.6 & 84.2 & -& -& 75.1\\
    DART \cite{Yang_2022_CVPR}     &  68.7 & 96.4 & 99.0  & 66.3 & 72.5 & 97.8 & 99.5 & 78.2   & 83.6 & -    & -    & 75.7 & 82.0 & -    & -    & 73.8    \\
    MRCN \citep{zhang2023mrcn} &  68.9 & 95.2 & 98.4 & 65.5 & 76.0 & 98.3 & 99.7 & 79.8 & 91.4 & 98.0 & 99.0 & 84.6 & 88.3 & 96.7 & 98.5 & 81.9     \\
    CAJ \cite{ye2021channel}     &  69.9 & 95.7 & 98.5 & 66.9 & 76.3 & 97.9 & 99.5 & 80.4  & 85.0 & 95.5 & 97.5 & 79.1 & 84.8 & 95.3 & 97.5 & 77.8  \\
    MPANet \citep{wu2021discover}    &  70.6 & 96.2 & 98.8 & 68.2 & 76.7 & 98.2 &  99.6 & 81.0 & 82.8 & -    & -    & 80.7 & 83.7 & -    & -    & 80.9    \\
    MMN \cite{zhang2021towards}  &  70.6 & 96.2 & 99.0 & 66.9 & 76.2 & 97.2 & 99.3 & 79.6 &   91.6 & 97.7 & \textcolor{blue}{98.9} & 84.1 & 87.5 & 96.0 & 98.1 & 80.5 \\ 
    DCLNet \citep{Sun_2022_ACM}   &  70.8 & -    & -    & 65.3 & 73.5 & -    & -    & 76.8    & 81.2 & -    & -    & 74.3 & 78.0 & -    & -    & 70.6    \\
    MAUM \citep{LiuJialun_2022_CVPR}&  71.7  & -    & - & 68.8  & 77.0  & -    & - & 81.9 & 87.9 & - & - & 85.1 & 87.0 & - & -& 84.3  \\
    DEEN \citep{zhang2023diverse}    & 74.7 & \textcolor{blue}{97.6} & \textcolor{blue}{99.2} & 71.8 & 80.3 & 99.0 & 99.8 & \textcolor{blue}{83.3}  & 91.1 & 97.8 & \textcolor{blue}{98.9} & 85.1 & 89.5 & 96.8 & \textcolor{blue}{98.4} & 83.4   \\
    MUN \citep{Yu_2023_ICCV} &  76.2 & \textcolor{red}{97.8} & - & \textcolor{blue}{73.8} & 79.4 & 98.1 & - & 82.1 & \textcolor{blue}{95.2} & \textcolor{blue}{98.9} & - & \textcolor{blue}{87.2} & \textcolor{blue}{91.9} & \textcolor{blue}{98.0} & - & 85.0\\
    MSCLNet \citep{zhang2022modality}    &  77.0 & \textcolor{blue}{97.6} & \textcolor{blue}{99.2} & 71.6 & 78.5 & \textcolor{blue}{99.3} & \textcolor{blue}{99.9} & 81.2 & 84.2 & -    & -    & 81.0 & 83.9 & -    & -    & 78.3\\  
    SGIEL \citep{Feng_2023_CVPR}     &  \textcolor{blue}{77.1} & 97.0 & 99.1 & 72.3 & \textcolor{blue}{82.1} & 97.4 & 98.9 & 83.0 & 92.2 & -    & -   & 86.6 & 91.1 & -    & -    & \textcolor{blue}{85.2} \\
    \hline\hline
    FDNM (Ours)             &  \textcolor{red}{77.8} & \textcolor{red}{97.8} & \textcolor{red}{99.6} & \textcolor{red}{75.1}   & \textcolor{red}{87.3} & \textcolor{red}{99.4} & \textcolor{red}{100.0} & \textcolor{red}{89.1} & \textcolor{red}{95.5} & \textcolor{red}{99.0} & \textcolor{red}{99.6} & \textcolor{red}{90.0} & \textcolor{red}{94.0} & \textcolor{red}{98.5} & \textcolor{red}{99.3} & \textcolor{red}{88.7}   \\
    \hline
    \end{tabular} 
    \vspace{-0.4cm}
\end{table*}

\subsection{Datasets and Evaluation Metrics}
The proposed FDNM method is evaluated on three challenging VIReID datasets, including SYSU-MM01 \citep{wu2017rgb}, RegDB \citep{nguyen2017person} and LLCM \citep{zhang2023diverse}. 
The SYSU-MM01 dataset \citep{wu2017rgb} includes $34,167$ images of 491 identities. The training set contains $22,258$ VIS images and $11,909$ IR images of $395$ identities, and the query set contains $3,803$ IR images of $96$ identities. Following \citep{ye2018hierarchical}, the gallery set includes the all-search mode and the indoor-search mode. 
The RegDB dataset \citep{nguyen2017person} contains $8,240$ images of 412 identities captured by a pair of overlapping VIS and IR cameras. Following \citep{ye2018hierarchical}, both VIS to IR and IR to VIS modes are used to evaluate the competing methods with a random half-half split of the training and testing set. %And the final results are based on an average of 10 times testing. %We repeat 10 trails with a random half-half split of the dataset: Half of identities are used for training and the other half are used for testing. The final results are based on an average of 10 times testing.
The LLCM dataset \citep{zhang2023diverse} is a low-light cross-modality dataset, which contains 46,767 images of 1,064 identities. The training set includes 30,921 images of 713 identities, and the testing set includes 13,909 images of 351 identities. Both the VIS to IR and IR to VIS modes are used to evaluate the proposed FDNM method. 

Following \cite{ye2021deep, zhang2023diverse}, the standard cumulative matching characteristics (CMC) and the mean average precision (mAP) are used as the evaluation metrics.% in this paper. 

\subsection{Implementation Details}

The proposed FDNM employs the two-stream ResNet50 network \citep{he2016deep, ye2020dynamic} as the backbone network, which is initialized with pre-trained weights on ImageNet \citep{Deng2009ImageNet}. Following \cite{zhang2021towards, zhang2023mrcn, tan2023exploring}, all input images are resized to $3 \times 384 \times 192$, and the commonly used random grayscale, random horizontal flip and random erasing \citep{zhong2020random} are adopted for data augmentation techniques. During the training phase, 4 VIS images and 4 IR images of 6 identities are randomly sampled to build a mini-batch. The SGD optimizer is adopted and the momentum is set to 0.9. The initial learning rate is set to $1\times10^{-2}$ and is linearly increased to $1\times10^{-1}$ with 10 epochs. Then, the learning rate is respectively decayed to $1\times10^{-2}$, $1\times10^{-3}$ and $1\times10^{-4}$ after the 20th, 80th and 120th epochs, until a total of 150 epochs. For the hyperparameter $\lambda_1$ in Eq. (\ref{eq9}), we experimentally set it to 1.0. %For the margin parameter in the triplet loss \cite{hermans2017defense}, we experimentally set it to 0.3. %For the coefficient $\lambda_1$ in Eq. \ref{eq1} and Eq. \ref{eq2}, we experimentally set it to 1.0. For the coefficient $\mu$ in Eq. (\ref{eq3}), we set it to 3. 
\vspace{-0.05cm}
\subsection{Comparison with State-of-the-Art Methods}
To verify the superiority of the proposed FDNM, we compare it with several state-of-the-art methods. The quantitative results on the SYSU-MM01 and RegDB datasets are shown in Tab. \ref{tab1}, and the results on the LLCM dataset are shown in Tab. \ref{tab2}. 

\textbf{SYSU-MM01:} From Tab. \ref{tab1}, we can observe that the proposed FDNM achieves the best performance among all the competing methods. Specifically, for the all-search mode, FDNM achieves 77.8\% in Rank-1 accuracy and 75.1\% in mAP, surpassing the SGIEL \citep{Feng_2023_CVPR} method by 0.7\% in Rank-1 accuracy and the MUN \citep{Yu_2023_ICCV} method by 1.3\% in mAP, respectively. For the indoor-search mode, FDNM achieves 87.3\% in Rank-1 accuracy and 89.1\% in mAP, which outperforms the SGIEL \cite{Feng_2023_CVPR} method by 5.2\% in Rank-1 accuracy and the DEEN \cite{zhang2023diverse} method by 5.8\% in mAP, respectively. The results show that the proposed FDNM is not only suitable for outdoor scenes, but also applicable for indoor scenes, exhibiting good generalization ability. % for the VIReID task.

\textbf{RegDB:} From Tab. \ref{tab1}, we can observe that the proposed FDNM achieves the best performance against the state-of-the-art methods. Specifically, for the VIS to IR mode, FDNM achieves 95.5\% in Rank-1 accuracy and 90.0\% in mAP, which outperforms the second-best method (MUN \cite{Yu_2023_ICCV}) by 2.8\% in mAP. 
For the IR to VIS mode, the proposed FDNM obtains 94.0\% in Rank-1 accuracy and 88.7\% in mAP, which outperforms the MUN \cite{Yu_2023_ICCV} method by 2.1\% in Rank-1 accuracy and the SGIRL \cite{Feng_2023_CVPR} method by 3.5\% in mAP, respectively. 
The results validate the effectiveness of our method. 
Moreover, the results also demonstrate that the proposed FDNM can effectively alleviate the modality discrepancy between the VIS and IR images.

\textbf{LLCM:} From Tab. \ref{tab2}, we can observe that the proposed FDNM achieves competitive performance on the LLCM dataset. Specifically, for the IR to VIS mode, FDNM achieves 56.6\% in Rank-1 accuracy and 62.7\% in mAP, respectively. For the VIS to IR mode, FDNM obtains 70.2\% in Rank-1 accuracy and 55.8\% in mAP, respectively. %The LLCM \citep{zhang2023diverse} dataset is a challenging low-light cross-modality dataset. 
Although DEEN \citep{zhang2023diverse} achieves an impressive performance through multi-branch networks and complex attention modules, it requires significant computational resources and training time. Compared with DEEN, the proposed FDNM is more effective and efficient for the VIReID task.

\begin{table}\small
  \centering
      \caption{Comparisons between the proposed FDNM and several state-of-the-art methods on the LLCM dataset.}% under single-shot setting. The best and the second-best results are bold and underlined marked, respectively.}
    \label{tab2}  \vspace{-0.2cm}
  \tabcolsep=0.07cm
\begin{tabular}{l|cccc|cccc}
\hline 
\multirow{2}{*}{Methods}     & \multicolumn{4}{c|}{IR to VIS}  & \multicolumn{4}{c}{VIS to IR}  \\ \cline{2-9} 
& \multicolumn{1}{c}{R-1} & \multicolumn{1}{c}{R-10} & \multicolumn{1}{c}{R-20} & \multicolumn{1}{c|}{mAP} & \multicolumn{1}{c}{R-1} & \multicolumn{1}{c}{R-10} & \multicolumn{1}{c}{R-20} & \multicolumn{1}{c}{mAP} \\ \hline
    DDAG \citep{ye2020dynamic}	   & 42.4 & 72.7 & 80.6 & 49.0 & 51.4 & 81.5 & 88.3 & 38.8  \\
    LbA	\citep{park2021learning}   & 42.8 & 77.4 & 86.1 & 51.0 & 54.8 & 85.1 & 91.6 & 40.8  \\
    AGW \citep{ye2021deep}	       & 49.1 & 79.1 & 85.9 & 55.8 & 63.7 & 88.7 & 92.8 & 47.2  \\
    CAJ  \citep{ye2021channel}	   & 49.9 & 78.9 & 85.8 & 56.4 & 63.7 & 88.0 & 92.4 & 47.7  \\
    MMN \citep{zhang2021towards}   & 50.1 & 79.8 & 87.3 & 56.7 & 64.0 & 88.7 & 93.1 & 48.5  \\
    MRCN \citep{zhang2023mrcn}	   & 51.3 & 80.1 & 87.2 & 57.7 & 65.3 & 88.1 & 93.1 & 49.5  \\
    DART \citep{Yang_2022_CVPR}	   & 53.0 & 80.8 & 87.1 & 59.3 & 65.3 & 89.4 & 93.3 & 51.1  \\
    DEEN \citep{zhang2023diverse}  & \textcolor{blue}{55.5} & \textcolor{blue}{83.9} & \textcolor{blue}{90.0} & \textcolor{blue}{62.1} & \textcolor{blue}{69.2} & \textcolor{blue}{91.0} & \textcolor{red}{95.1} & \textcolor{blue}{55.5}  \\\hline\hline
    FDNM (Ours)                   & \textcolor{red}{56.6} & \textcolor{red}{84.1} & \textcolor{red}{90.4} & \textcolor{red}{62.7} & \textcolor{red}{70.2} & \textcolor{red}{91.1} & \textcolor{blue}{94.8} & \textcolor{red}{55.8}      \\    \hline 
    \end{tabular}\vspace{-0.2cm}
\end{table}

\begin{table}[t]\small
  \centering
\caption{Ablation studies for different components of the proposed FDNM on the SYSU-MM01 dataset.}  \label{tab3} \vspace{-0.2cm}
  \tabcolsep=0.08cm  
 \fontsize{9pt}{10pt}\selectfont
\begin{tabular}{cc|ccc|cc|cc}
\hline 
\multicolumn{5}{c|}{Settings}    & \multicolumn{2}{c|}{All Search}  & \multicolumn{2}{c}{Indoor Search}  \\ \hline
\multicolumn{1}{c}{Base.}      &\multicolumn{1}{c|}{Local}      &\multicolumn{1}{c}{AGP}      &\multicolumn{1}{c}{ANM}      &\multicolumn{1}{c|}{$L_{cnm}$}      & \multicolumn{1}{c}{R-1}  & \multicolumn{1}{c|}{mAP} & \multicolumn{1}{c}{R-1}  & \multicolumn{1}{c}{mAP} \\ \hline

\checkmark &  &             &          &     & 68.7 & 66.0 &76.1  & 80.2 \\
 & \checkmark &             &          &     & 69.5 & 66.8 & 77.3  & 81.7  \\
\checkmark & \checkmark &             &          &         & 71.9 & 68.3 & 81.7 & 83.7   \\ \hline
\checkmark & \checkmark & \checkmark  &           & & 74.6 & 71.7 & 84.2 & 87.1      \\ 
\checkmark & \checkmark &  & \checkmark &  &   73.2 & 70.8 & 82.7 & 85.5  \\ 
%\checkmark & \checkmark & \checkmark  & \checkmark  &     \\ 
\checkmark & \checkmark &  & \checkmark & \checkmark       & \textcolor{blue}{76.0} & \textcolor{blue}{73.6} & \textcolor{blue}{86.6} & \textcolor{blue}{88.3} \\ 
\checkmark & \checkmark & \checkmark  & \checkmark & \checkmark  & \textcolor{red}{77.8}  & \textcolor{red}{75.1}   & \textcolor{red}{87.3} & \textcolor{red}{89.1}  \\
 \hline
    \end{tabular}%}
\vspace{-0.45cm}
\end{table}

\subsection{Ablation Studies}

\textbf{Ablation studies for the effectiveness of different components.} To demonstrate the contribution of each component in FDNM, we conduct some ablation studies by evaluating different components on the SYSU-MM01 dataset. As we can see in Tab. \ref{tab3}, the ``Base." and ``Local" represent the two-stream ResNet50 network \citep{he2016deep, ye2020dynamic} with only global features or local features. The experiments show that: (1) the proposed AGP module can enhance the performance of the baseline by 2.7\% in Rank-1 accuracy and 3.4\% in mAP, which indicates the proposed AGP module can effectively alleviate the VIS-IR modality discrepancy by utilizing the key information in amplitude component to guide the phase component for learning discriminative feature representations. (2) The proposed ANM module slightly improves the performance of the baseline, but the results are not satisfactory. This is because directly applying the IN layers to feature maps may corrupt some identity information and prevent the learning of discriminative feature representations. By contrast, the proposed center-guided nuances mining loss $\mathcal{L}_{cnm}$ can greatly improve the performance of the ANM module. This shows that the $\mathcal{L}_{cnm}$ loss can effectively mine the cross-modality nuances of the amplitude features while reducing the modality differences between the VIS and IR images. (3) By integrating the AGP module, the ANM module and the $\mathcal{L}_{cnm}$ loss into an end-to-end frequency domain learning framework, the proposed FDNM effectively alleviates the modality discrepancy, resulting in a desirable performance.
 
\begin{table}[t]\small
  \centering
      \caption{Ablation studies with which block of ResNet-50 to integrate the proposed AGP module on the SYSU-MM01 dataset.}
    \label{tab4}
  \tabcolsep=0.33cm  \vspace{-0.2cm}
\begin{tabular}{l|cc|cc}
\hline
\multirow{2}{*}{Settings} & \multicolumn{2}{c|}{All Search}& \multicolumn{2}{c}{Indoor Search} \\    \cline{2-5}
& \multicolumn{1}{c}{R-1}  & \multicolumn{1}{c|}{mAP} & \multicolumn{1}{c}{R-1}  & \multicolumn{1}{c}{mAP} \\ \hline
    baseline             & 71.9  & 68.3 & 81.7   & 83.7     \\ \hline
    AGP after block-0    & 72.3  & 69.1 & 82.1   & 84.5  \\
    AGP after block-1	 & \textcolor{red}{74.1}  & \textcolor{red}{71.3} & \textcolor{red}{83.8}   & \textcolor{red}{86.4}      \\ 
    AGP after block-2	 & \textcolor{blue}{73.1}  & \textcolor{blue}{70.9} & \textcolor{blue}{82.9}   & \textcolor{blue}{86.0}   \\
    AGP after block-3 	 & 71.1  & 68.5 & 80.2  & 82.5 \\ 
    AGP after block-4    & 70.8  & 67.2 & 79.5  & 81.7  \\\hline 
    \end{tabular}\vspace{-0.2cm}
\end{table}

\begin{figure}[t]% \vspace{-0.15cm}
\centerline{\includegraphics[height=3.4cm,width=8.4cm]{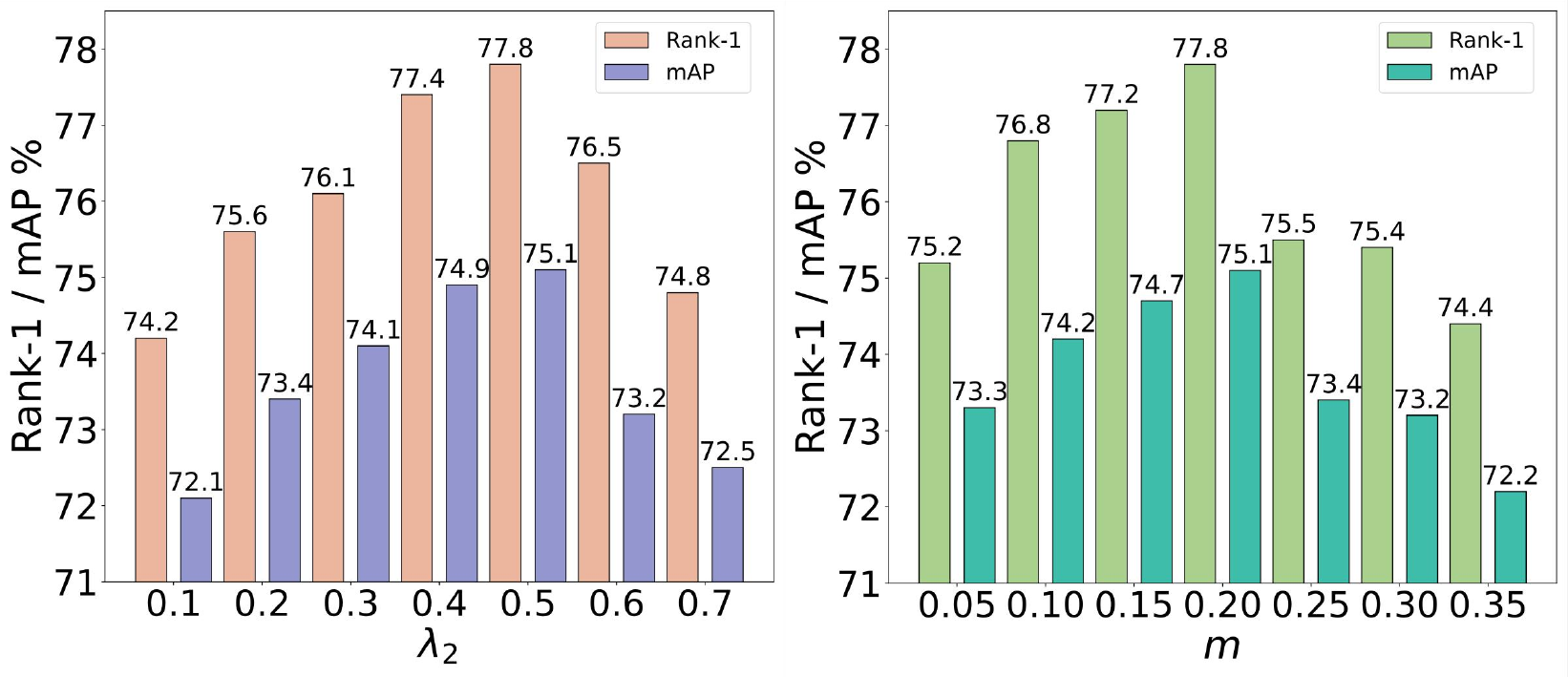}} \vspace{-0.35cm}
\caption{Comparisons of different $\lambda_2$ and $m$ values on the SYSU-MM01 dataset.}
\label{img5}
\vspace{-0.55cm}
\end{figure}

\begin{figure*}[t]
\centerline{\includegraphics[height=6.1cm,width=17.1cm]{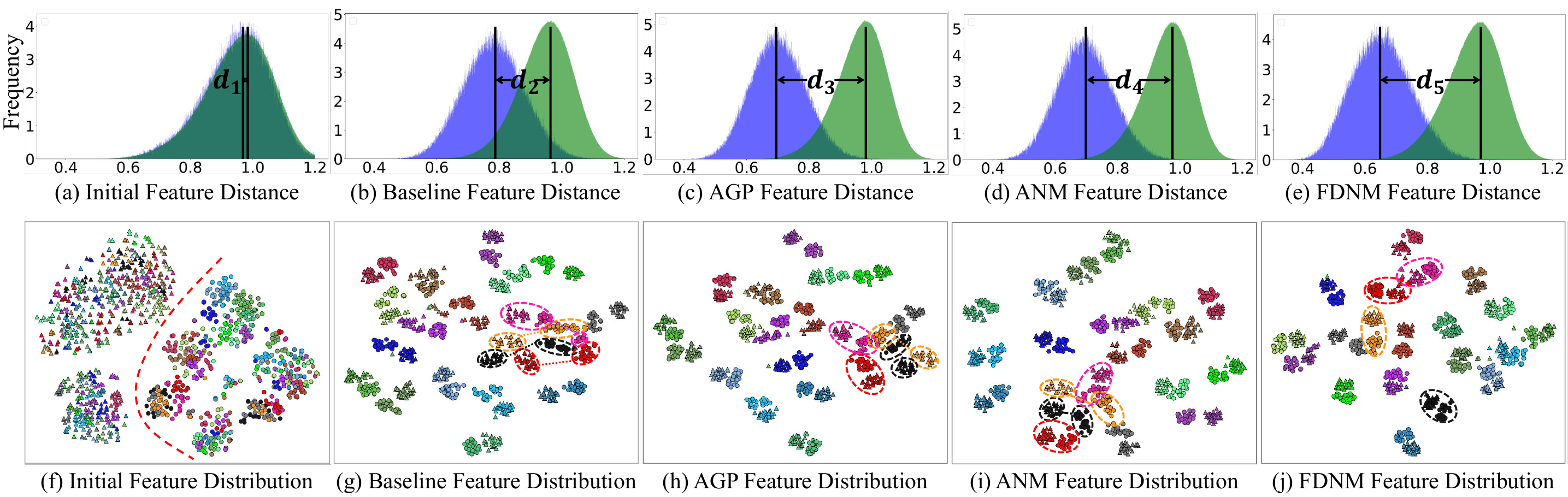}} \vspace{-0.22cm}
\caption{(a-e) The distribution of the cosine distance between VIS-IR intra-class features and inter-class features, which are indicated by blue and green colors, respectively. (f-j) The t-SNE \citep{laurens2008Visualizing} distribution of the VIS-IR features. Different colors represent different identities. The “circle” and “triangle” markers represent the VIS and IR features, respectively.}
\label{img6}
 \vspace{-0.42cm}
\end{figure*}

%\textbf{The influence of which stage of ResNet-50 to plug the AGP module.} The proposed AGP module is a plug-and-play module which can be plugged after any stage of the backbone network. In our experiments, we use ResNet-50 as the backbone, which has five stages: stage-0 to stage-4. We plug AGP module after different stages to study how different stages will affect the performance of the AGP module. As shown in Tab. \ref{tab4}, when AGP is plugged after stage-0 to stage-3, the performance gradually increases, which shows the salient information in amplitude component can effectively guide the phase component to learn discriminative feature representations. However, when AGP is plugged after stage-4 and stage-5, the performance drops rapidly, which is beacuse AGP module after stage-4 and stage-5 will damage discriminative information in phase component. Based on the above analysis, to balance model performance and efficiency, we plug the proposed AGP module into stage-1 and stage-2 after the backbone if not specified.

\textbf{Ablation studies with which block to integrate the AGP module.} The proposed AGP module is a plug-and-play module and it can be integrated into any blocks of the backbone network. In our experiments, we utilize the two-stream ResNet-50 as the backbone, which includes 5 residual blocks: block-0 to block-4. To investigate the influence of integrating the proposed AGP module, we integrate it at different blocks. As shown in Tab. \ref{tab4}, when the AGP module is integrated after block-0 to block-2, the performance improves, suggesting that the key information in the amplitude component can effectively guide the phase component to learn discriminative feature representations. However, when the AGP module is integrated after block-3 and block-4, the performance drops rapidly. This is because integrating the AGP module after block-3 and block-4 may damage some discriminative phase information. Based on the above results, to balance the performance and efficiency, we integrate the proposed AGP module after the block-1 and block-2 of the backbone model.

\textbf{The influence of the hyperparameters $\lambda_2$ and $m$.} %In Eq. \ref{eq7} and Eq. \ref{eq9}, we introduce two hyperparameters $\lambda_2$ and $m$ to balance the relative importance of different loss terms. 
%To evaluate the influence of $\lambda_2$ and $m$, we conducte quantitative comparisons and present the results in Fig. \ref{img5}. As can be seen, the best performance is achieved when $\lambda_2$ is set to 0.8 and $m$ is set to 0.2, respectively. The proposed FDNM adopts a triplet-like strategy to mine diverse nuances in the amplitude component and alleviate the modality discrepancies between the VIS and IR images. With the increase of $\lambda_2$ and $m$ values, the performance of the FDNM can gradually improve. However, beyond a certain threshold, $\lambda_2$ and $m$ may cause the FDNM to collapse. Therefore, we set $\lambda_2$ is set to 0.8 and $m$ is set to 0.2 in our experiments. 
The proposed center-guided nuances mining loss $\mathcal{L}_{cnm}$ utilizes two hyperparameters $\lambda_2$ and $m$ with a triplet-like strategy to mine diverse nuances in the amplitude component. To evaluate the influence of $\lambda_2$ and $m$, we conducte quantitative comparisons and present the results in Fig. \ref{img5}. As can be seen, with the increase of $\lambda_2$ and $m$ values, the performance of FDNM gradually improves. However, when the values of $\lambda_2$ and $m$ are more than a certain threshold, the performance of FDNM may collapse. Therefore, the values of $\lambda_2$ and $m$ are set to 0.5 and 0.2, respectively.% in our experiments. 

\begin{table}\small
\vspace{0.09cm}
  \centering
      \caption{Evaluation of the proposed FDNM on two visible-infrared face recognition datasets. The results of Rank-1 accuracy and false acceptance rate (F:\%) are reported. }
    \label{tab6}
  \tabcolsep=0.105cm
  \vspace{-0.2cm}
\begin{tabular}{l|ccc|ccc}  
\hline 
\multirow{2}{*}{Methods}     & \multicolumn{3}{c|}{Oulu}  & \multicolumn{3}{c}{BUAA}  \\ \cline{2-7} 
& \multicolumn{1}{c}{R-1} & \multicolumn{1}{c}{F:1$\%$} & \multicolumn{1}{c|}{F:0.1$\%$} & \multicolumn{1}{c}{R-1} & \multicolumn{1}{c}{F:1$\%$} & \multicolumn{1}{c}{F:0.1$\%$} \\ \hline
    IDR \citep{He2017Learning}	   & 94.3 & 73.4 & 46.2 & 94.3 & 93.4 & 84.7  \\
    ADFL	\citep{Wu2018Coupled}   & \textcolor{blue}{95.5} & 83.0 & 60.7 & 95.2 & 95.3 & 88.0  \\
    %VSA \cite{yu2021lamp} & \textcolor{blue}{99.9} & 96.8 & 82.3 & 98.0 & 98.2 & 92.5 \\
    PACH  \citep{Duan_2020_CVPR}	   & \textcolor{red}{100} & 97.9 & 88.2 & \textcolor{blue}{98.6} & \textcolor{blue}{98.0} & 93.5  \\
    CAJ \citep{ye2021channel}   & \textcolor{red}{100} & 97.9 & 87.0 & 98.0 & 97.7 & 93.9  \\
    TOPLight \citep{Yu_2023_CVPR}	   & \textcolor{red}{100} & \textcolor{blue}{98.9} & \textcolor{blue}{91.7} & 98.3 & \textcolor{blue}{98.0} & \textcolor{blue}{94.6}  \\ \hline\hline
    LightCNN-29 \citep{wu2018light}	   & \textcolor{red}{100} & 97.9 & 87.0 & 98.0 & 97.7 & 93.7  \\

    FDNM (Ours)                   & \textcolor{red}{100} & \textcolor{red}{99.1} & \textcolor{red}{92.2} & \textcolor{red}{99.8} & \textcolor{red}{98.9} & \textcolor{red}{95.5}      \\    \hline 
    \end{tabular}
        \vspace{-0.5cm}
\end{table}

\subsection{Visualization Analysis}
\textbf{Intra-class and inter-class distances.} To evaluate the effectiveness of the proposed method, we analyze the frequency distribution of the intra-class and inter-class distances between the VIS and IR features on SYSU-MM01. As shown in Fig. \ref{img6} (a-e). Compared the initial feature distances (Fig. \ref{img6} (a)) and baseline feature distances (Fig. \ref{img6} (b)) with the feature distances obtained by the proposed method Fig. \ref{img6} (c-e), the means (i.e., the vertical lines) of intra-class and inter-class distances are pushed away by using the proposed AGP module, ANM module and FDNM ($d_5$ \textgreater $d_3$ \textgreater $d_2$ \textgreater $d_1 \approx 0$, and $d_5$ \textgreater $d_4$ \textgreater $d_2$ \textgreater $d_1 \approx 0$). This result indicates that the proposed method can effectively alleviate the modality discrepancy between the VIS and IR images, thereby achieving better performance. 

\textbf{Feature distribution.} For further validating the effectiveness of the proposed FDNM, we visualize the feature distribution utilizing t-SNE \citep{laurens2008Visualizing} on the SYSU-MM01 dataset. As shown in Fig. \ref{img6} (f-j), a total of 24 persons are randomly selected from the testing set of the SYSU-MM01 dataset. Compared with the initial feature distribution in Fig. \ref{img6} (f) and the baseline's feature distribution in Fig. \ref{img6} (g), the feature distribution of the proposed AGP module, ANM module and FDNM can enlarge the distances between the features of different identities, and shorten the distances between the features from the same identities, which effectively reduce the VIS-IR modality discrepancies.

\subsection{Visible-Infrared Face Recognition}
To investigate the generalizability of the proposed FDNM, we evaluate it on two VIS-IR face recognition datasets (i.e., Oulu-CASIA NIR-VIS \cite{JieCVPR2009} and BUAA-VisNir \cite{huang2012buaa}). Similar to \cite{ye2021channel, Yu_2023_CVPR}, we also adopt the LightCNN-29 \cite{wu2018light, Duan_2020_CVPR} as the backbone network for the VIS-IR face recognition task. The training details are exactly the same as \cite{ye2021channel, Yu_2023_CVPR}. The results are shown in Tab. \ref{tab6}. As can be seen, the proposed FDNM can also enhance the performance of the VIS-IR face recognition task, which further indicates the effectiveness and generalization of the proposed FDNM.

\section{Conclusion}

In this paper, we aim to mine cross-modality frequency nuances for the VIReID task and propose a novel frequency domain nuances mining (FDNM) method. The proposed FDNM includes an amplitude guided phase module and an amplitude nuances mining module. 
These two modules are mutually beneficial to jointly reduce the VIS-IR modality discrepancy. Besides, we propose a center-guided nuances mining loss to encourage the ANM module to preserve discriminative identity information while discovering diverse cross-modality nuances. %To the best of our knowledge, this is the first work that explores the potential frequency information for VIReID research. 
Extensive experiments on three challenging VIReID datasets demonstrate the superior performance of the proposed FDNM method. 
Besides, we also validate the effectiveness and generalization of our method on the VIS-IR face recognition task.

{
    \small
    \bibliographystyle{ieeenat_fullname}
    \bibliography{main}
}


\clearpage
\setcounter{page}{1}
\maketitlesupplementary

\section{Retrieval results}
\label{Retrieval}

To further evaluate the proposed FDNM, we compare the retrieval results obtained by our method with those obtained by the baseline method on several VIS-IR image pairs from the SYSU-MM01 \cite{wu2017rgb} (under the multi-shot setting and the all-search mode), RegDB \cite{nguyen2017person} and LLCM datasets \cite{zhang2023diverse}. 
For the RegDB and LLCM datasets, both the VIS to IR and IR to VIS modes are evaluated. 
The results are shown in Fig. \ref{img8}. For each retrieval case, the query images shown in the first column are given images, and the gallery images shown in the following columns are retrieved images obtained by the baseline method and the proposed FDNM method. The retrieved images with the green bounding boxes belong to the same identities as the query images, while the images with the red bounding boxes are opposite to the query images. 
The experiments show that: 

(1) In general, the proposed FDNM method can effectively improve the ranking results with more green bounding boxes ranked in the top positions on those three challenging VIReID datasets, which demonstrates that the proposed FDNM can effectively alleviate the modality discrepancy between the VIS and IR images and improve the performance of the VIReID task.

(2) The proposed FDNM is not only suitable for the VIS and NIR (near infrared) match tasks such as for the SYSU-MM01 and LLCM datasets, but also applicable for the VIS and TIR (thermal infrared) match tasks such as for the RegDB dataset, exhibiting good generalization ability. 

(3) The ranking results obtained by the proposed FDNM method on the LLCM dataset with more green bounding boxes ranked in the top positions, which indicates that the proposed FDNM method also has good robustness and generalization to extremely low-light environments at night.

%\textbf{The effectiveness on different baseline methods.} To further demonstrate the effectiveness of the proposed FDNM on different baseline methods, we conduct ablation studies on some common baseline (DDAG \cite{ye2020dynamic}, ResNet-50 \citep{he2016deep} and AGW \citep{ye2020dynamic}). The overall settings remain the same on these baseline methods. Tab. \ref{tab7} shows the obtained results. As we can see, the proposed FDNM can improve performance under different baseline methods, indicating that FDNM can be applied to different baseline methods to reduce the modality discrepancy between the VIS and IR images by exploring and mining the potential frequency nuances for the VIReID task. 

\end{document}